\documentclass[10pt,twocolumn,letterpaper]{article}

\usepackage{iccv}
\usepackage{times}
\usepackage{epsfig}
\usepackage{graphicx}
\usepackage{amsmath}
\usepackage{amssymb}
\usepackage{color}
\usepackage{times}
\usepackage{epsfig}
\usepackage{pifont}
\usepackage{cite}
\usepackage{subfigure}
\usepackage{multirow}
\usepackage{diagbox}
\usepackage{rotating}
\usepackage{booktabs}
\usepackage{colortbl}
\usepackage{overpic}
\usepackage[table]{xcolor}
\usepackage{textcomp}
\usepackage{contour}
\usepackage[breaklinks=true,colorlinks,citecolor=blue,urlcolor=blue,bookmarks=false]{hyperref}
\newcommand{\ConfInf}{\vspace{-.7in} {\normalsize \normalfont \color{blue}{
   IEEE International Conference on Computer Vision (ICCV) 2019}} \vspace{.45in} \\}

\iccvfinalcopy 
\ificcvfinal\fi

\def\iccvPaperID{1596} 

\ifdefined \GramaCheck
  \newcommand{\CheckRmv}[1]{}
  \newcommand{\figref}[1]{Figure 1}%
  \newcommand{\tabref}[1]{Table 1}%
  \newcommand{\secref}[1]{Section 1}
  \newcommand{\equref}[1]{Equation 1}
\else
  \newcommand{\CheckRmv}[1]{#1}
  \newcommand{\figref}[1]{Fig.~\ref{#1}}%
  \newcommand{\tabref}[1]{Tab.~\ref{#1}}%
  \newcommand{\secref}[1]{Sec.~\ref{#1}}
  \newcommand{\equref}[1]{Eq.~(\ref{#1})}
\fi

\newcolumntype{x}[1]{>{\centering\arraybackslash\hspace{0pt}}p{#1}}

\newlength\savedwidth
\newcolumntype{"}{@{\hskip\tabcolsep\vrule width 0.6pt\hskip\tabcolsep}}
\newcommand{\whline}[1]{\noalign{\global\savedwidth\arrayrulewidth \global\arrayrulewidth #1}%
	\hline \noalign{\global\arrayrulewidth\savedwidth}}

\def\etal{{\em et al.~}}

\graphicspath{{Imgs/}}

\begin{document}

\title{\ConfInf EGNet: Edge Guidance Network for Salient Object Detection}

\author{Jia-Xing Zhao, Jiang-Jiang Liu, Deng-Ping Fan, Yang Cao, Ju-Feng Yang, 
	Ming-Ming Cheng\thanks{M.M. Cheng (cmm@nankai.edu.cn) is the corresponding author.}\\
    TKLNDST, CS, Nankai University \\
{\tt\small http://mmcheng.net/egnet/}
}

\maketitle
\thispagestyle{empty}

\begin{abstract}
   Fully convolutional neural networks (FCNs) have shown  their advantages in the salient object detection task.  
	However, most existing FCNs-based methods still suffer from coarse object boundaries. 
   In this paper, to solve this problem, we focus on the  complementarity between salient edge information and salient object information.
   Accordingly,   
   we present an edge guidance network (EGNet)  for salient object detection with three steps to simultaneously model 
  these two kinds of complementary information in a single network. 
   In the first step, we extract the salient object features by a progressive fusion way. 
   In the second step, 
   we integrate the local edge information and global location information to
   obtain the salient edge features.       
   Finally, to sufficiently leverage these complementary features,  
   we couple the same salient edge features with salient object features at various resolutions.
   Benefiting from the rich edge information and location information in salient edge features, the fused 
   features can help locate salient objects, especially their boundaries more accurately.  
   Experimental results demonstrate that   
   the proposed method performs favorably
   against the state-of-the-art methods on six widely used datasets without any pre-processing 
   and post-processing. 
   The source code is available at \url{http://mmcheng.net/egnet/}. 
\end{abstract}

\vspace{-4pt}
\section{Introduction} \label{introduction}
\vspace{-4pt}
The goal of salient object detection (SOD) is to find the most visually distinctive objects in an image. It has received widespread attention recently 
and been widely used in many vision and image processing related areas, 
such as content-aware image editing \cite{cheng2010repfinder},
object recognition \cite{rutishauser2004bottom},
photosynth \cite{tog09Sketch2Photo},
non-photo-realist rendering \cite{rosin2013artistic}, weakly supervised semantic segmentation \cite{hou2018selferasing} and
image retrieval \cite{he2012mobile}. 
Besides, there are many works focusing on video salient object detection 
\cite{fan2019shifting, RANet2019} and RGB-D salient object detection 
\cite{fan2019rethinking, zhao2019contrast}. 

\newcommand{\addFig}[1]{\includegraphics[width=0.24\linewidth]{itro/#1}}
\newcommand{\addFigsd}[1]{\addFig{#1.jpg} &\addFig{#1_1.png} & \addFig{#1_0.png} & \addFig{#1.png}}
\CheckRmv{
\begin{figure}
	\centering
	\footnotesize
	\setlength\tabcolsep{0.2mm}
	\renewcommand\arraystretch{0.5}
	\begin{tabular}{cccc}
		\addFigsd{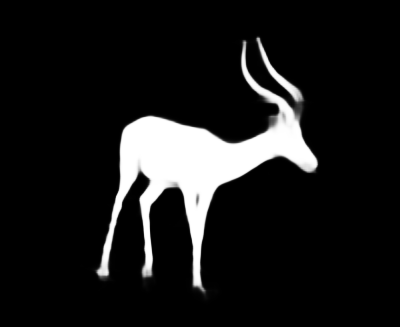}\\
		\addFigsd{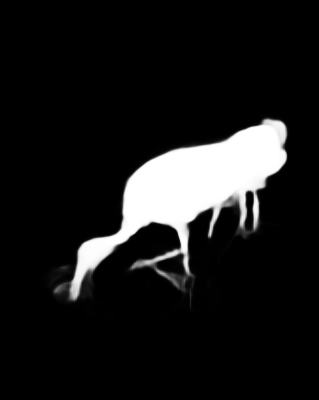}\\
		Source& Baseline & Ours& GT 
		\\
	\end{tabular}
	\vspace{1mm}
	\caption{Visual examples of our method. After we model and fuse the salient edge information, the salient object boundaries 
		become clearer.
	} 
	\label{fig:intro}
	\vspace{-15pt}
\end{figure}
}

Inspired by cognitive studies of visual attention 
\cite{itti2001computational, parkhurst2002modeling, einhauser2003does},
early works are mainly based on the fact that contrast plays the most
important role in saliency detection.
These methods benefit mostly from either global or local 
contrast cues and their learned fusion weights.
Unfortunately, these hand-crafted features though can locate
the most salient objects sometimes, the produced saliency maps are
with irregular shapes because of the undesirable segmentation methods and
unreliable when the contrast between the foreground and the background is inadequate.

Recently, convolutional neural networks (CNNs) \cite{lecun1998gradient}
have successfully broken the limits of traditional hand-crafted features,
especially after the emerging of Fully Convolutional
Neural Networks (FCNs) \cite{long2015fully}.
These CNN-based methods have greatly refreshed the leaderboards on almost all
the widely used benchmarks and are gradually replacing conventional
salient object detection methods because of the efficiency as well as
high performance.
%
In SOD approaches based 
on CNNs architecture, the majority of them which regard the image patches \cite{zhao2018flic, ZhaoCvm2018flic} as
input use the multi-scale or multi-context information to obtain the final saliency 
map. 
Since the fully convolutional network is proposed for pixel labeling problems, 
several end-to-end deep architectures \cite{HouPami19Dss, HouCvpr2017Dss,li2016deep, jia2019richer, li2019deep, zhang2019salient, zhao2019optimizing, wangiccv2019lfsd} for salient object detection appear. 
The basic unit of output saliency map becomes per pixel from the image region.
On the one hand, the result highlights the details because each pixel has its saliency 
value. 
However, on the other hand, it ignores the structure information which is 
important for  SOD.
%

With the increase of the network receptive field, the positioning of salient objects becomes more 
and more accurate. However, at the same time, spatial coherence is also ignored. 
Recently, to obtain the fine edge details, some SOD U-Net \cite{ronneberger2015u}  based works \cite{liu2016dhsnet, zhang2017amulet, zhang2018bi,liu2018picanet} 
used a bi-directional or recursive way to refine the high-level features with the local information. 
However, the boundaries of salient objects are still not explicitly modeled. 
The complementarity between the salient edge information and salient object information has not been noticed.
Besides, there are some methods using pre-processing (Superpixel) \cite{hu2017deep} or 
post-processing (CRF) \cite{li2016deep,HouCvpr2017Dss,liu2018picanet} to preserve the object boundaries. 
The main inconvenience with these approaches is their low inference speed. 

In this paper, we focus on the complementarity between salient edge information and salient object information. 
We aim to leverage the salient edge features to help the salient object features locate objects, especially their boundaries more accurately. In summary, this paper makes three major contributions:
\begin{itemize}
	\vspace{-4pt}
    \item We propose an EGNet to explicitly model complementary 
    salient object information and salient edge information within 
    the network to preserve the salient object boundaries. 
    At the same time, the salient edge features are 
    also helpful for localization. 
    \vspace{-4pt} 
    \item  Our model jointly optimizes these two complementary tasks by allowing them to mutually help each other, which significantly improves the predicted saliency maps.
    \vspace{-4pt}
    \item We compare the proposed methods with 15 state-of-the-art approaches 
    on six widely used datasets. 
    Without bells and whistles, our 
    method achieves the best performance under three evaluation metrics. 
\end{itemize}

\CheckRmv{
\begin{figure*}[tp]
  \centering
  \includegraphics[width=.95\linewidth]{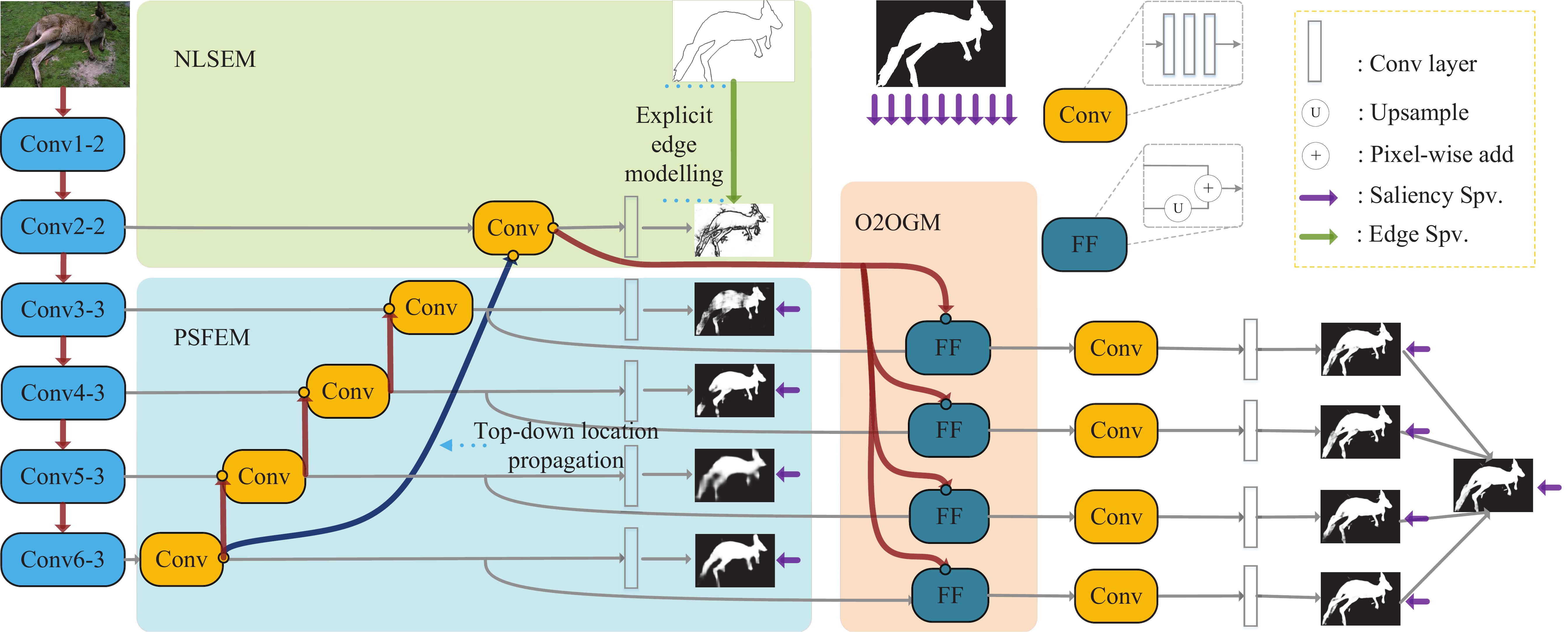}
  \caption{The pipeline of the proposed approach. 
  	We use brown thick lines to represent information flows between the scales. 
   PSFEM: progressive salient object features extraction module. NLSEM: non-local salient edge features extraction module. 
   O2OGM : one-to-one guidance module. FF: feature fusion. Spv.: supervision.
  }\label{fig:arch}
\end{figure*}
}

\vspace{-10pt}
\section{Related Works}
\vspace{-4pt}

Over the past years, some methods were proposed to detect the salient objects 
in an image. 
Early methods predicted the saliency map using a bottom-up pattern by the hand-craft feature, such as 
contrast \cite{cheng2015global}, boundary background \cite{yang2013saliency, zhu2014saliency}, 
center prior \cite{jiang2013salient, klein2011center} and so on \cite{itti1998model, jiang2013salient, WangECCV2016ksr}. 
More details are introduced in \cite{borji2015salient,BorjiCVM2019, fan2018salient}.

Recently, Convolutional neural networks (CNNs) perform their advantages and refresh the state-of-the-art records in many 
fields of computer vision. 
%

%
Li \etal \cite{li2015visual} resized the image regions to three different scales  to extract the multi-scale features and then 
aggregated these multiple saliency maps to obtain the final prediction map. 
Wang \etal \cite{wang2015deep} designed a neural network to extract the local estimation for the input patches and integrated these 
features with the global contrast and geometric information to describe the image patches. 
%
However, the result is limited by the performance of image patches in these methods. 
In \cite{long2015fully}, long \etal firstly proposed a network (FCN) to predict the semantic label for each pixel. 
Inspired by FCN, more and more pixel-wise saliency detection methods were proposed. 
Wang \etal \cite{wang2016saliency} proposed a recurrent FCN architecture for salient object detection. 
%
%
Hou \etal proposed a short connection \cite{HouCvpr2017Dss,HouPami19Dss} based on HED \cite{xie2015holistically} to 
integrate the low-level features and high-level features to solve the scale-space problem.
In \cite{zhang2017learning}, Zhang \etal introduced a reformulated dropout and an effective hybrid upsampling to 
learn deep uncertain convolutional features to encourage robustness and accuracy. 
In \cite{zhang2017amulet}, Zhang \etal explicitly aggregated the multi-level features into multiple resolutions and 
then combined these feature maps by a bidirectional aggregation method. 
Zhang \etal \cite{zhang2018bi} proposed a bi-directional message-passing model to integrate multi-level 
features for salient object detection. 
Wang \etal \cite{wang2018salient} leveraged the fixation maps to help the model to locate the salient object more accurately. 
In \cite{luo2017non}, Luo \etal proposed a U-Net based architecture which contains an IOU edge loss to 
leverage the edge cues to detect the salient objects. 
In other saliency-related tasks, some methods of using edge cues have appeared. 
In \cite{li2017instance}, li \etal generated the contour of the object to obtain the salient instance segmentation results. 
In \cite{li2018contour}, li \etal leveraged the well-trained contour detection models to generate the saliency masks
to overcome the limitation caused by manual annotations.

%
%
Compared with most of the SOD U-Net based methods \cite{liu2016dhsnet, zhang2017amulet, zhang2018bi,liu2018picanet}, 
we explicitly model edge information within the network to leverage the edge cues. 
Compared with the methods which use the edge cues \cite{guan2018edge,zhuge2018boundary,zhang2017deep}, 
the major differences are that we use a single base network and jointly optimize the salient edge 
detection and the salient object detection, allowing them to help each other mutually. which results in better performance. 
Compared with NLDF \cite{luo2017non}, they implemented a loss function inspired by the Mumford-Shah
function \cite{mumford1989optimal} to penalize errors on the edges. 
Since the salient edges are derived from salient objects through a fixed sober operator, this penalty essentially
only affects the gradient in the neighborhood of salient edges on feature maps. 
In this way, the edge details are optimized to some extent, but the complementarity between salient edge detection and 
salient object detection is not sufficiently utilized. 
In our method, we design two modules to extract these two kinds of features independently. Then 
we fuse these complementary features by a one-to-one guidance module. In this way, the 
salient edge information can not only improve the quality of edges but also make the localization more accurate. 
The experimental part verifies our statement.

\section{Salient Edge Guidance Network}
The overall architecture is shown in \figref{fig:arch}. 
In this section, we begin by describing the motivations in \secref{sec_arc}, then 
introduce the adopted salient object feature extraction module and the proposed non-local salient edge features extraction module in
\secref{sec_edge}, and 
finally introduce the proposed one-to-one guidance module 
in \secref{sec_guide}.

\vspace{-4pt}
\subsection{Motivation} \label{sec_arc}
\vspace{-4pt}
The pixel-wise salient object detection methods have shown their advantages compared with region-based methods. 
However, they ignored the spatial coherence in the images, resulting in the unsatisfied salient object boundaries.
Most methods \cite{HouCvpr2017Dss,HouPami19Dss,liu2018picanet,li2019deep,zhang2017amulet,zhang2018bi} hope to solve this problem by fusing multi-scale information. Some methods \cite{HouCvpr2017Dss,liu2018picanet,li2016deep} used the post-processing such as CRF to refine the salient object boundaries. In NLDF \cite{luo2017non}, they proposed an IOU loss to affect the gradient of the location 
around the edge. 
None of them pay attention to the complementarity between salient edge detection and salient object detection. 
A good salient edge detection result can help salient object detection task in both segmentation and localization, and vice versa. 
Based on this idea, we proposed an EGNet to model and fuse the complementary salient edge information and salient object information within 
a single network in an end-to-end manner. 


\vspace{-4pt}
\subsection{Complementary information modeling} \label{sec_edge}
\vspace{-4pt}
Our proposed network is independent of the backbone network. 
Here we use the VGG network 
suggested by other deep learning based methods \cite{HouCvpr2017Dss, luo2017non} to describe the proposed method. 
First, we truncate the last three fully connected layers. 
Following DSS \cite{HouCvpr2017Dss,HouPami19Dss}, we connect another side path to the last pooling layer in VGG. 
Thus from the backbone network, we obtain six side  features Conv1-2, Conv2-2, Conv3-3, Conv4-3, Conv5-3, Conv6-3. 
Because the Conv1-2 is too close to the input and the receptive field is too small, we throw away this side path
$S^{(1)}$. There are five side paths $S^{(2)}, S^{(3)}, S^{(4)}, S^{(5)}, S^{(6)}$ remaining in our method. 
For simplicity, these five features could be denoted by a backbone features set $C$: 
\vspace{-4pt}
\begin{equation}
    C = \{C^{(2)}, C^{(3)}, C^{(4)}, C^{(5)}, C^{(6)}\}, 
\end{equation}
where $C^{(2)}$ denotes the Conv2-2 features and so on. 
Conv2-2 preserves 
better edge information \cite{zhang2017amulet}. 
Thus we leverage the $S^{(2)}$ to extract the  edge features and other side paths to extract the salient object features.  

\vspace{-4pt}
\subsubsection{Progressive salient object features extraction}
\vspace{-4pt}
\quad As shown in PSFEM of \figref{fig:arch}, to obtain richer context features, we leverage the 
widely used architecture U-Net \cite{ronneberger2015u} to generate the multi-resolution features. 
Different from the original U-Net, in order to obtain more robust salient object features, We add three convolutional layers  (Conv in \figref{fig:arch}) on each side path, and 
after each convolutional layer, a ReLU layer is added to ensure the nonlinearity. To illustrate simply, we use the $T$ (\tabref{details}) to denote these 
convolutional layers and ReLU layers. 
Besides, deep supervision is used on each side path. We adopt a convolutional layer to convert the feature maps to the single-channel 
prediction mask and use $D$ (\tabref{details}) to denote it. 
The details of the convolutional layers could be found in \tabref{details}.

\CheckRmv{
\begin{table}[t]
	\centering
	\scriptsize
	\renewcommand{\arraystretch}{1}
	\renewcommand{\tabcolsep}{2.2mm}
	\begin{tabular}{l|c|c|c|c|c|c|c|c|c|c|c|c}
		\whline{0.5pt}
		S&\multicolumn{3}{c|}{$T_1$} & \multicolumn{3}{c|}{$T_2$}  & \multicolumn{3}{c|}{$T_3$} & \multicolumn{3}{c}{$D$} \\
		\whline{0.5pt}
		2& 3 & 1& 128 & 3 & 1& 128 & 3 & 1 & 128  &3 & 1& 1\\
		3& 3 & 1& 256 & 3 & 1& 256 & 3 & 1 & 256  &3 & 1& 1 \\
		4& 5 & 2& 512 & 5 & 2& 512 & 5 & 2 & 512  &3 & 1& 1 \\
		5& 5 & 2& 512 & 5 & 2& 512 & 5 & 2 & 512  &3 & 1& 1\\
		6& 7 & 3& 512 & 7 & 3& 512 & 7 & 3 & 512  &3 & 1& 1\\
		\whline{0.5pt}
	\end{tabular}
	\vspace{1mm}
	\caption{Details of each side output. T denotes the feature enhance module (Conv shown 
		in \figref{fig:arch}). Each T contains three convolutional layers: $T_1$, $T_2$, $T_3$ and 
		three followed ReLu layers. We show the kernel size, padding and channel number of each convolutional layer. 
		For example, 3, 1, 128 denote a convolutional layer whose kernel size is 3, padding is 1, channel number
		is 128.  D denotes the transition layer which converts the 
		multi-channel feature map to one-channel activation map. S denotes the side path.} \label{details}
\end{table}
}

\vspace{-4pt}
\subsubsection{Non-local salient edge features extraction}
\vspace{-4pt}

\quad In this module, we aim to model the salient edge information and extract the salient edge features. 
As mentioned above, the  Conv2-2 preserves better edge information. Hence we extract local edge information from Conv2-2. 
However, in order to get salient edge features, only local information is not enough. 
High-level semantic information or location information is also needed. 
When information is progressively returned from the top level to the low level like the U-Net architecture, the high-level location information is gradually diluted.
Besides, 
the receptive field of the top-level is the largest, and the location is the most accurate. 
Thus we design a top-down location propagation to propagate the top-level location information to the side path $S^{(2)}$ 
to  restrain the non-salient edge. 
The fused features $\bar{C}^{(2)}$ could be denoted as:
\vspace{-4pt}
\begin{equation}
    \bar{C}^{(2)} = C^{(2)} + \mathit{Up}(\phi(\mathit{Trans}(\hat{F}^{(6)}; \theta)); C^{(2)}),
\end{equation}
where $\mathit{Trans}(*; \theta)$ is a convolutional layer with parameter $\theta$, which aims to change the number of channels of the feature, and $\phi()$ denotes a ReLU activation function. $Up(*; C^{(2)})$ is bilinear interpolation operation which aims to up-sample * to 
the same size as $C^{(2)}$. 
On the right of the equation, the second term denotes the features from the higher side path. 
To illustrate clearly, we use $\mathit{UpT(\hat{F}^{(i)};\theta, C^{(j)})}$ to represent 
$\mathit{Up}(\phi(\mathit{Trans}(\hat{F}^{(i)}; \theta)); C^{(j)})$. 
$\hat{F}^{(6)}$ denotes the enhanced features in side path $S^{(6)}$. 
The enhanced features $\hat{F}^{(6)}$ could be represented as $\mathnormal{f}(C^{(6)}; W_T^{(6)})$, and 
the enhanced features in $S^{(3)}, S^{(4)}, S^{(5)}$ could be computed as: 
\vspace{-4pt}
\begin{equation}\label{ehance_com}
    \hat{F}^{(i)} = \mathnormal{f}(C^{(i)}+\mathit{UpT}(\hat{F}^{(i + 1)};\theta, C^{(i)}) ;W_T^{(i)}),
\end{equation}
where $W_T^{(i)}$ denotes the parameters in $T^{(i)}$ and 
$f(*; W_T^{(i)})$ denotes a series of convolutional and non-linear operations with parameters $W_T^{(i)}$. 

After obtaining the guided features $\bar{C}^{(2)}$, similar with other side paths, we 
add a series convolutional layers to enhance the guided feature, then the final \textit{ salient edge features }
${F_E}$ in  $S^{(2)}$ could be computed 
as $\mathnormal{f}(\bar{C}^{(2)}; W_T^{(2)})$. The configuration details could be found in \tabref{details}.
To model the salient edge feature explicitly, 
we add an extra salient edge supervision to supervise the salient edge features.
We use the cross-entropy loss which could be defined as: 
\begin{gather}
    \mathcal{L}^{(2)}(F_{E}; W_D^{(2)}) =  - \sum_{j\in Z_+}{\log Pr( \mathit{y}_j = 1 |  F_E; W_D^{(2)})} \notag \\ 
    - \sum_{j\in Z_-}{\log Pr( \mathit{y}_j = 0 | F_E; W_D^{(2)})},  
\end{gather}
where $Z_+$ and $Z_-$ denote the salient edge pixels set and background pixels set, respectively. 
$W_D$ 
denotes the parameters of the transition layer as shown in \tabref{details}. 
$Pr( \mathit{y}_j = 1 | F_E; W_D^{(2)})$ is the prediction map in which each value denotes the salient edge confidence for 
the pixel.
In addition, the supervision added on the salient object detection side path can be represented as:
\begin{gather}
    \mathcal{L}^{(i)}(\hat{F}^{(i)}; W_D^{(i)}) =  -\sum_{j\in Y_+}{\log Pr( \mathit{y}_j = 1 | \hat{F^{(i)}}; W_D^{(i)})} \notag \\ 
    - \sum_{j\in Y_-}{\log Pr( \mathit{y}_j = 0 | \hat{F}^{(i)}; W_D^{(i)})}, ~ i \in [3,6], \label{cross-entropy}
\end{gather}
where $Y_+$ and $Y_-$ denote the salient region pixels set and non-salient pixels set, respectively. 
Thus the total loss $\mathbb{L} $ in the complementary information modeling could be denoted as:
\vspace{-4pt}
\begin{equation}
\mathbb{L} = \mathcal{L}^{(2)}(F_E; W_D^{(2)}) +  \sum_{i = 3}^{6}\mathcal{L}^{(i)}(\hat{F}^{(i)}; W_D^{(i)}).     
\end{equation}
 \vspace{-4pt}

\vspace{-4pt}
\subsection{One-to-one guidance module} \label{sec_guide}
\vspace{-4pt}
After obtaining the complementary salient edge features and salient object features, 
we aim to leverage the salient edge features to guide the salient object features to perform better on both segmentation 
and localization.  The simple way is to 
fuse the $F_E$ and the $\hat{F}^{(3)}$. 
It will be better to sufficiently leverage the multi-resolution salient object features. 
However, the disadvantage of fusing the salient edge features and multi-resolution salient object features progressively from down to top is that salient edge features are diluted when salient object features are fused. 
Besides, the goal is to fuse  salient object features and salient edge features to utilize complementary information 
to obtain better prediction results. 
Hence, we propose a one-to-one guidance module. Moreover, experimental parts validate our view.

Specifically,  we add sub-side paths for $S^{(3)}$, $S^{(4)}$, $S^{(5)}$, $S^{(6)}$. 
In each sub-side path, by fusing the salient edge features into enhanced salient object features, 
we make the location of high-level predictions more accurate, and more importantly, the segmentation details become better.  
The  salient edge guidance features (s-features) could be denoted as: 
\vspace{-4pt}
\begin{equation}
    {G}^{(i)} = \mathit{UpT}(\hat{F}^{(i)};\theta, F_E) + F_E,  i \in [3,6].
\end{equation}
Then similar to the PSFEM, we adopt a series of convolutional layers $T$ in each sub-side path to further enhance 
the s-features and a transition layer $D$ to convert the multi-channel feature map to one-channel prediction map. 
Here in order to illustrate clearly, we denote the $T$ and $D$ as  $T'$ and $D'$ in this module. 
By \equref{ehance_com}, we obtain the enhanced s-features $\hat{G}^{(i)}$. 

Here we also add deep supervision for these enhanced s-features.
For each sub-side output prediction map, the loss can be calculated as: 
\begin{gather}
    \mathcal{L}^{(i)'}(\hat{G}^{(i)}; W_{D'}^{(i)}) =  -\sum_{j\in Y_+}{\log Pr( \mathit{y}_j = 1 | \hat{G}^{(i)}; W_{D'}^{(i)})} \notag \\ 
    - \sum_{j\in Y_-}{\log Pr( \mathit{y}_j = 0 | \hat{G}^{(i)}; W_{D'}^{(i)})}, ~ i \in [3,6].
\end{gather}
Then we fuse the multi-scale refined prediction maps to obtain a fused map. The loss function 
for the fused map can be denoted as: 
\vspace{-4pt}
\begin{equation}
    \mathcal{L}_f'(\hat{G}; W_{D'}) =  \sigma (Y, \sum_{i=3}^{6}{\beta_i f(\hat{G}^{(i)}; W_{D'}^{(i)})}), 
\end{equation}
where the $\sigma (*, *)$ represents the cross-entropy loss between prediction map and saliency ground-truth, which 
has the same form to \equref{cross-entropy}. 
Thus the  loss for this part  and the total for the proposed network could be expressed as:
\vspace{-4pt}
\begin{equation}\label{e23}
\begin{split}
&\mathbb{L}' = \mathcal{L}_f'(\hat{G}; W_{D'}) + \sum_{i=3}^{i=6} { \mathcal{L}^{(i)'}(\hat{G}^{(i)}; W_{D'}^{(i)})} \\
&\mathbb{L}_t = \mathbb{L} + \mathbb{L}'. \\
\end{split}
\end{equation}
\vspace{-4pt}

\CheckRmv{
\begin{table*}[tp!] 
	\centering
	\scriptsize
	\renewcommand{\arraystretch}{1.2}
	\newcommand{\trb}[1]{\textcolor{red}{\textbf{#1}}}
	\newcommand{\tgb}[1]{\textcolor{green}{\textbf{#1}}}
	\newcommand{\tbb}[1]{\textcolor{blue}{\textbf{#1}}}
	\renewcommand{\tabcolsep}{0.8mm}
	\begin{tabular}{l "ccc " ccc " ccc " ccc " ccc " ccc}
		\whline{1pt}
		& \multicolumn{3}{c"}{ECSSD \cite{yan2013hierarchical}} & \multicolumn{3}{c"}{PASCAL-S \cite{li2014secrets}} & \multicolumn{3}{c"}{DUT-O \cite{yang2013saliency}} & \multicolumn{3}{c"}{HKU-IS \cite{li2015visual}} & \multicolumn{3}{c"}{SOD \cite{martin2001database,movahedi2010design}} & \multicolumn{3}{c}{DUTS-TE \cite{wang2017learning}} \\ 
		  & MaxF~$\uparrow$ & MAE~$\downarrow$ &S~$\uparrow$  & MaxF~$\uparrow$ & MAE~$\downarrow$ &S~$\uparrow$ & MaxF~$\uparrow$ & MAE~$\downarrow$ &S~$\uparrow$ & MaxF~$\uparrow$ & MAE~$\downarrow$ &S~$\uparrow$ & MaxF~$\uparrow$ & MAE~$\downarrow$  &S~$\uparrow$ & MaxF~$\uparrow$ & MAE~$\downarrow$ &S~$\uparrow$ \\
		\hline
		\multicolumn{19}{c}{VGG-based} \\
		\hline
		\textbf{DCL$^*$}~\cite{li2016deep} & 0.896 & 0.080 & 0.863 & 0.805 & 0.115& 0.791  & 0.733 & 0.094 & 0.743 & 0.893 & 0.063 & 0.859 & 0.831 & 0.131 & 0.748 & 0.786 & 0.081 & 0.785\\
		\textbf{DSS$^*$}~\cite{HouCvpr2017Dss,HouPami19Dss} &0.906 & 0.064& 0.882 & 0.821 & 0.101&0.796  & 0.760 & 0.074 &0.765& 0.900 & 0.050 &0.878 & 0.834 & 0.125&0.744 & 0.813 & 0.065 & 0.812 \\
		\textbf{MSR}~\cite{li2017instance} & 0.903 & 0.059&0.875 & 0.839 & 0.083&0.802 & 0.790 & 0.073&0.767 & 0.907 & 0.043&0.852 & 0.841 & 0.111 &0.757& 0.824 & 0.062&0.809 \\
		\textbf{NLDF}~\cite{luo2017non} &  0.903 & 0.065 & 0.875 & 0.822 & 0.098 & 0.803 & 0.753 & 0.079 & 0.750 & 0.902 & 0.048&0.878 & 0.837 & 0.123 & 0.756 & 0.816 & 0.065 & 0.805\\
		\textbf{RAS}~\cite{chen2018reverse} &  0.915 & 0.060& 0.886 & 0.830 &0.102&0.798&0.784&0.063 & 0.792 
		& 0.910&0.047&0.884&0.844&0.130&0.760 & 0.800&0.060&0.827\\
		\textbf{ELD$^*$}~\cite{lee2016deep} & 0.865 & 0.082 & 0.839 & 0.772 & 0.122 &0.757 & 0.738 & 0.093 & 0.743 & 0.843 & 0.072&0.823 & 0.762 & 0.154 &0.705 & 0.747 & 0.092 & 0.749 \\
		\textbf{DHS}~\cite{liu2016dhsnet} & 0.905 & 0.062 & 0.884& 0.825 & 0.092 & 0.807 & - & - & -& 0.892 & 0.052 &0.869 & 0.823 & 0.128 &0.750 & 0.815 & 0.065 & 0.809\\
		\textbf{RFCN$^*$}~\cite{wangsaliency} &  0.898 & 0.097 & 0852 & 0.827 & 0.118 &0.799& 0.747 & 0.094&0.752 & 0.895 & 0.079 &0.860 & 0.805 & 0.161&0.730 & 0.786 & 0.090 &0.784\\
		\textbf{UCF}~\cite{zhang2017learning} & 0.908 & 0.080 & 0.884&0.820 & 0.127 &0.806& 0.735 & 0.131&0.748 & 0.888 & 0.073&0.874 & 0.798 & 0.164 &0.762& 0.771 & 0.116&0.777 \\	
		\textbf{Amulet}~\cite{zhang2017amulet} & 0.911 & 0.062 &0.894 & 0.826 & 0.092&0.820 & 0.737 & 0.083 &0.771& 0.889 & 0.052 &0.886& 0.799 & 0.146&0.753 & 0.773 & 0.075&0.796 \\
		\textbf{C2S}~\cite{li2018contour} & 0.909 & 0.057&0.891& 0.845 & 0.081&0.839 & 0.759 & 0.072 &0.783& 0.897 & 0.047 &0.886& 0.821 & 0.122&0.763 & 0.811 & 0.062&0.822\\
		\textbf{PAGR}~\cite{zhang2018progressive} & 0.924 & 0.064&0.889 & 0.847 & 0.089&0.818 & 0.771 & 0.071 &0.751& 0.919 & 0.047 & 0.889 & 0.841 & 0.146 &0.716& 0.854 & 0.055 &0.825 \\
		\hline
		\textbf{Ours }  & \tbb{0.941} & \tgb{0.044} &\tgb{0.913}& \tgb{0.863} & \tgb{0.076}&\tgb{0.848} & \tbb{0.826} & \tbb{0.056}&\tbb{0.813} & \tbb{0.929} & \tbb{0.034}&\tbb{0.910} & \tbb{0.869} & \tgb{0.110}&\tgb{0.788} & \tbb{0.880} & \tbb{0.043}&\tbb{0.866} \\
		\hline
		\multicolumn{19}{c}{ResNet-based } \\
		\hline
		\textbf{SRM$^*$}~\cite{wang2017stagewise} & 0.916 & 0.056&0.895 & 0.838 & 0.084&0.832 & 0.769 & 0.069&0.777 & 0.906 & 0.046&0.887 & 0.840 & 0.126&0.742 & 0.826 & 0.058&0.824 \\
		\textbf{DGRL}~\cite{wang2018detect} &0.921 & \tbb{0.043} &0.906& 0.844 & \tbb{0.075} &0.839& 0.774 & \tgb{0.062} &0.791& 0.910 & \tgb{0.036} &0.896& 0.843 & \tbb{0.103}&0.774 & 0.828 & \tgb{0.049} &0.836\\
		\textbf{PiCANet$^*$}~\cite{liu2018picanet} & \tgb{0.932} & 0.048&\tbb{0.914}  & \tbb{0.864} & {0.077}&\tbb{0.850} & \tgb{0.820} & 0.064&\tgb{0.808} & \tgb{0.920} & 0.044&\tgb{0.905} & \tgb{0.861} & \tbb{0.103}&\tbb{0.790} & \tgb{0.863} & 0.050&\tgb{0.850} \\
		\hline	
		\textbf{Ours } & \trb{0.943} & \trb{0.041} &\trb{0.918} & \trb{0.869} & \trb{0.074} & \trb{0.852} & \trb{0.842} & \trb{0.052} & \trb{0.818} & \trb{0.937} & \trb{0.031} &\trb{0.918} & \trb{0.890} & \trb{0.097}& \trb{0.807} & \trb{0.893} & \trb{0.039} & \trb{0.875}\\
		\whline{1pt}
	\end{tabular}
	\vspace{1mm}
	\caption{Quantitative comparison including max F-measure, MAE, and S-measure over six widely used datasets. 
		`-' denotes that corresponding methods are trained on that dataset. 
		$\uparrow \& \downarrow$ denote
		larger and smaller is better, respectively. 
		$^*$ means methods using pre-processing or post-processing.
		The best three results are marked
		in \textcolor{red}{red}, \textcolor{blue}{blue},
		and \textcolor{green}{green}, respectively. Our method
		achieves the state-of-the-art on these six widely used datasets under three evaluation metrics. 
	}\label{tab:results}
\end{table*}
}

\section{Experiments}

\subsection{Implementation Details}
\vspace{-4pt}
We train our model on DUTS \cite{wang2017learning} dataset followed by \cite{liu2018picanet, zhang2018bi, wang2017stagewise,zhang2018progressive}.
For a fair comparison, we use VGG \cite{simonyan2014very} and ResNet \cite{he2016deep} as backbone networks, respectively.    
Our model is implemented in PyTorch. 
All the weights of newly added convolution 
layers are initialized randomly with a truncated normal
($\sigma$ = 0.01), and the biases are initialized to 0. 
The hyper-parameters are set as followed: 
learning rate = 5e-5, weight decay = 0.0005, momentum = 0.9, 
loss weight for each side output is equal to 1. 
A back propagation is processing for each of the ten images. 
We do not use the validation dataset during training.
We train our model 24 epochs and divide the learning rate by 
10 after 15 epochs. During inference, we are able to obtain a predicted salient edge map and 
a set of saliency maps. 
In our method, 
we directly use the fused prediction map as the final saliency map.

\vspace{-4pt}
\subsection{Datasets and Evaluation Metric}
\vspace{-4pt}
We have evaluated the proposed architecture on six widely used public benchmark datasets: 
ECSSD \cite{yan2013hierarchical}, PASCAL-S \cite{li2014secrets}, DUT-OMRON \cite{yang2013saliency}, 
SOD \cite{martin2001database, jiang2013salient}, HKU-IS \cite{li2015visual}, DUTS \cite{wang2017learning}. 
ECSSD \cite{yan2013hierarchical} contains 1000 meaningful semantic images with various complex scenes. 
PASCAL-S \cite{li2014secrets} contains 850 images which are chosen from the validation set of 
the PASCAL VOC segmentation dataset \cite{everingham2010pascal}.  DUT-OMRON \cite{yang2013saliency} contains 5168 high-quality but challenging images. 
Images in this dataset contain one or more salient objects with a relatively complex background. 
SOD \cite{martin2001database} contains 300 images and is proposed for image segmentation. Pixel-wise annotations of 
salient objects are 
generated by \cite{jiang2013salient}.  It is one of the most challenging datasets currently. 
HKU-IS \cite{li2015visual} contains 4447 images with high-quality annotations, many of which have multiple disconnected 
salient objects. This dataset is split into 2500 training images, 500 validation images and 2000 test images. 
DUTS \cite{wang2017learning} is the largest salient object detection benchmark. It contains 10553 images for training and 
5019 images for testing. Most images are challenging with various locations and scales. 
Following most recent works \cite{liu2018picanet,wang2017stagewise,wang2018detect}, we use the DUTS dataset to train the proposed model. 

We use three widely used and standard metrics, F-measure, mean absolute error (MAE) \cite{borji2015salient}, and a recently proposed structure-based metric, namely 
S-measure \cite{fan2017structure}, to evaluate our model and other state-of-the-art models. 
F-measure is a harmonic mean of
average precision and average recall,
formulated as:
\vspace{-4pt}
\begin{equation}\label{equ:Fb_measure}
\begin{aligned}
    F_{\beta} &=\frac{(1 + \beta^2)Precision \times Recall}{\beta^2 \times Precision + Recall},
\end{aligned}
\end{equation}
we set $\beta^2=0.3$ to weigh
precision more than recall as suggested in~\cite{cheng2015global}. 
Precision denotes the ratio of detected salient pixels in the predicted saliency map. 
Recall denotes the ratio of detected salient pixels in the ground-truth map. 
Precision and recall are computed on binary images. 
Thus we should threshold the prediction map to binary map first. 
There are different precision and recall of different thresholds. 
We could plot the precision-recall curve at different thresholds.
Here we use the code provided by \cite{HouCvpr2017Dss,HouPami19Dss} for evaluation.  
Following most salient object detection methods \cite{HouCvpr2017Dss,HouPami19Dss, liu2016dhsnet, zhang2018bi}, 
we report the maximum F-measure from all precision-recall pairs.

MAE is a metric which evaluates the average difference between prediction map and ground-truth map. 
Let $P$ and $Y$ denote the saliency
map and the ground truth
that is normalized to [0, 1]. 
We compute the MAE score by:
\vspace{-4pt}
\begin{equation}\label{equ:MAE}
    \varepsilon = \frac{1}{W \times H} \sum_{x=1}^{W}\sum_{y=1}^{H} | P(x,y) - Y(x,y)|,
\end{equation}
where $W$ and $H$ are the width and
height of images, respectively.

S-measure focuses on evaluating the structural information of saliency maps, which is 
closer to the human visual system than F-measure. Thus we include S-measure for a more 
comprehensive evaluation. S-measure could be computed as:
\vspace{-4pt}
\begin{equation}\label{equ:S}
\mathtt{S} = \gamma \mathtt{S}_o + ( 1 - \gamma) \mathtt{S}_r, 
\end{equation}
where $\mathtt{S}_o$ and $\mathtt{S}_r$ denotes the region-aware and object-aware structural similarity and 
$\gamma$ is set as 0.5 by default. More details can be found in \cite{fan2017structure}.

\CheckRmv{
\begin{table}[tp!] 
	\centering
	\scriptsize
	\renewcommand{\arraystretch}{1.2}
	\renewcommand{\tabcolsep}{0.5mm}
	\begin{tabular}{l| c|c|c | c|c|c  }
		\hline
		\multirow{2}{*}{Model}
		& \multicolumn{3}{c|}{SOD} & \multicolumn{3}{c}{DUTS} \\
		\cline{2-7}
	& MaxF~$\uparrow$ & MAE~$\downarrow$ &S~$\uparrow$& MaxF~$\uparrow$ & MAE~$\downarrow$ &S~$\uparrow$\\
		\hline
		1. B& .851 & .116 & .780& .855 & .060 & .844\\
		2. B + edge\_PROG & .873 & .105 & .799& .872 & .051 & .851  \\
		3. B + edge\_TDLP& .882 & .100 & .807& .879 & .044 & .866 \\
		4. B + edge\_NLDF & .857 & .112 &.794& .866 & .053 & .860\\
		\whline{0.5pt}
		5. B + edge\_TDLP + MRF\_PROG & .882 &.106 &.796& .880 & .046 & .869\\
		6. B + edge\_TDLP + MRF\_OTO& .890 & .097& .807& .893 & .039 & .875\\
		\hline
	\end{tabular}
	\vspace{1mm}
	\caption{Ablation analyses on SOD \cite{martin2001database} and DUTS-TE \cite{wang2017learning}. Here, B denotes the baseline model. 
		edge\_PROG, edge\_TDLF, edge\_NLDF, MRF\_PROG, MRF\_OTO are introduced in the \secref{sec:ablation}.
	} \label{tab:module}
\end{table}
}

\subsection{Ablation Experiments and Analyses}\label{sec:ablation}
\vspace{-4pt}
In this section, with the DUTS-TR \cite{wang2017learning} as the training set, 
we explore the effect of different components in the proposed network 
over the  relatively difficult dataset SOD \cite{martin2001database} and the recently proposed big dataset DUTS-TE 
\cite{wang2017learning}.

\renewcommand{\addFig}[1]{\includegraphics[width=0.3\linewidth]{prs/#1.pdf}}
\CheckRmv{
\begin{figure*}
	\centering
	\footnotesize
	\setlength\tabcolsep{1.4mm}
	\begin{tabular}{cccc}
		\addFig{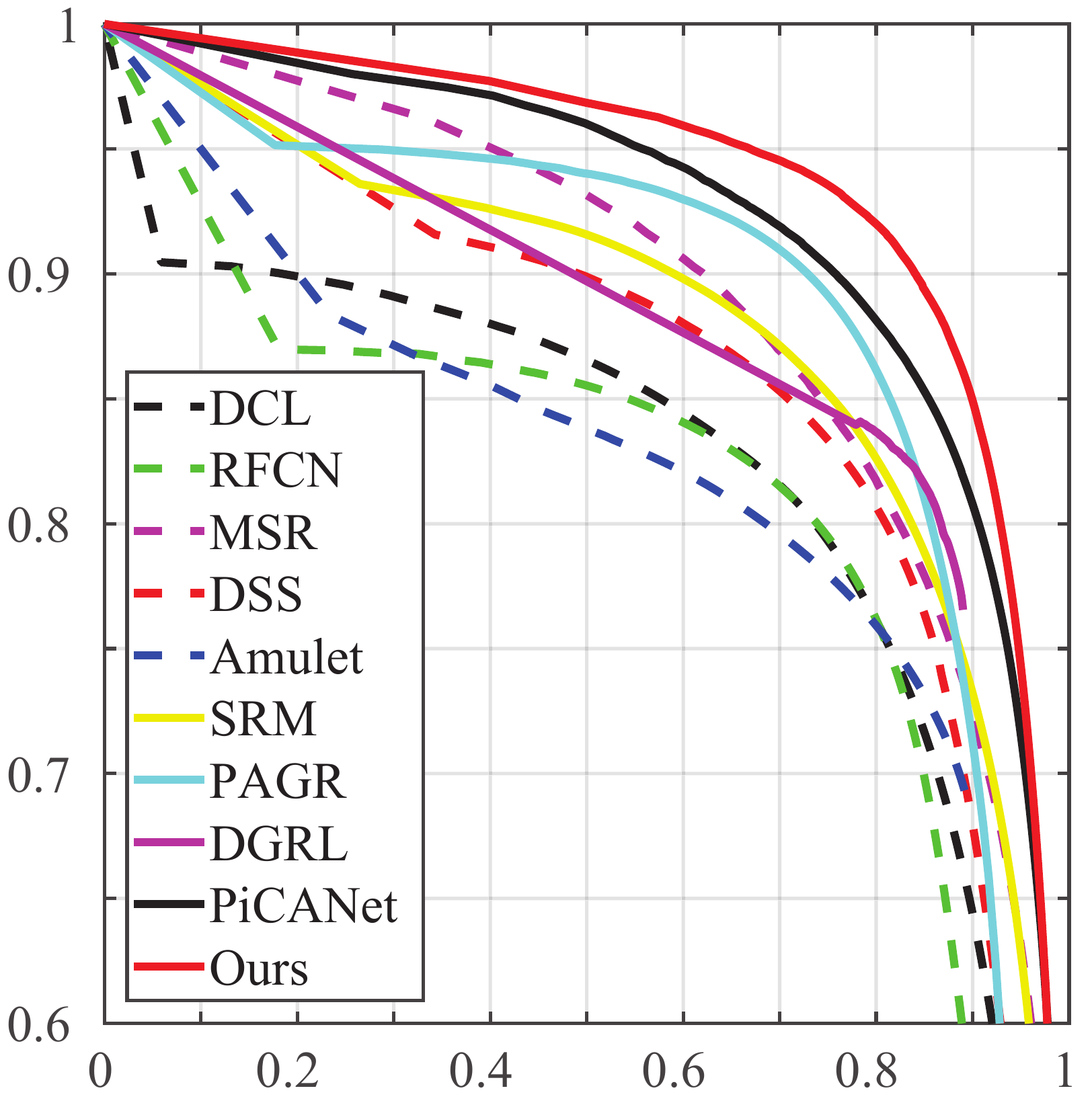} & \includegraphics[width=0.3035\linewidth]{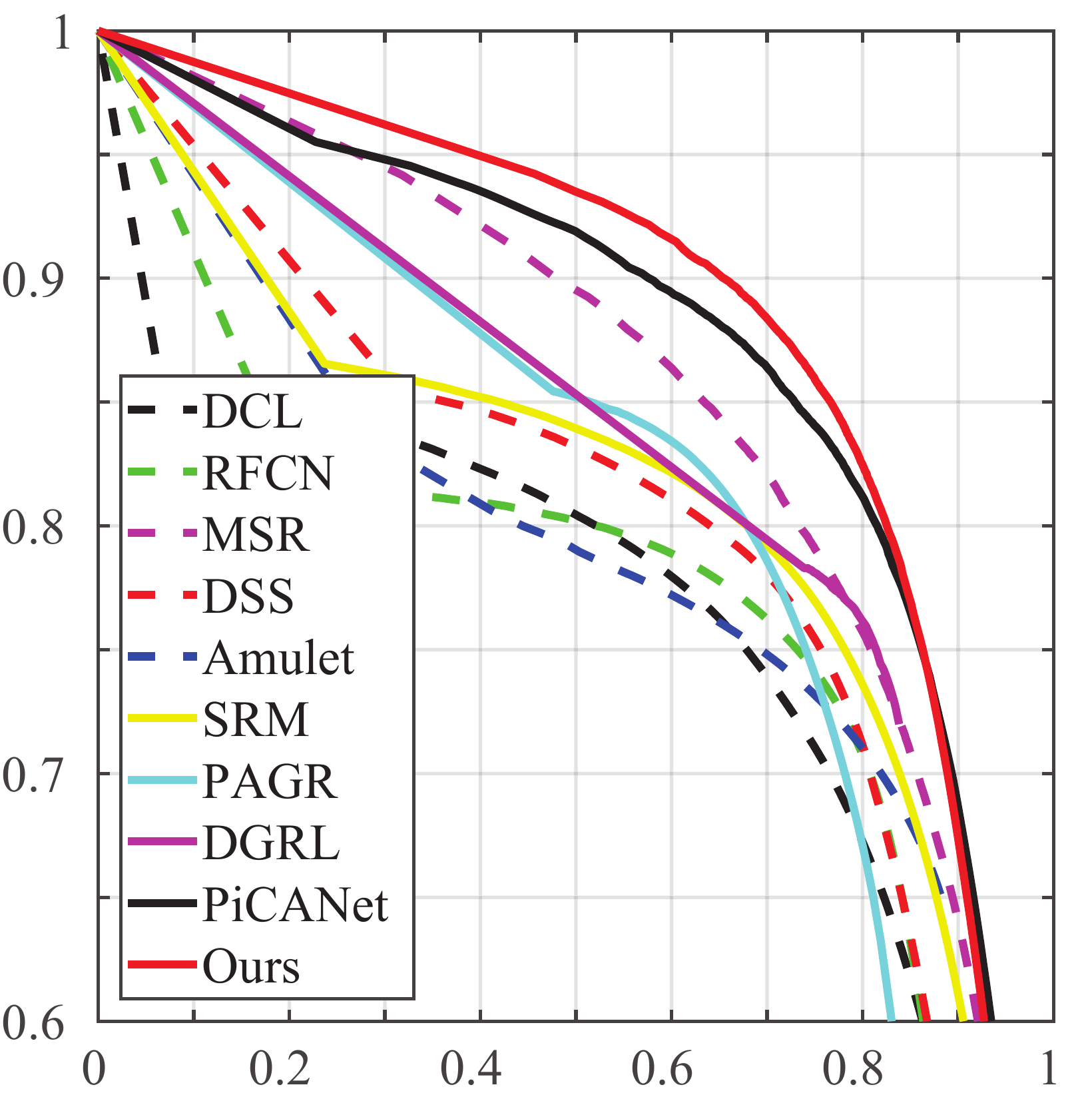} & \addFig{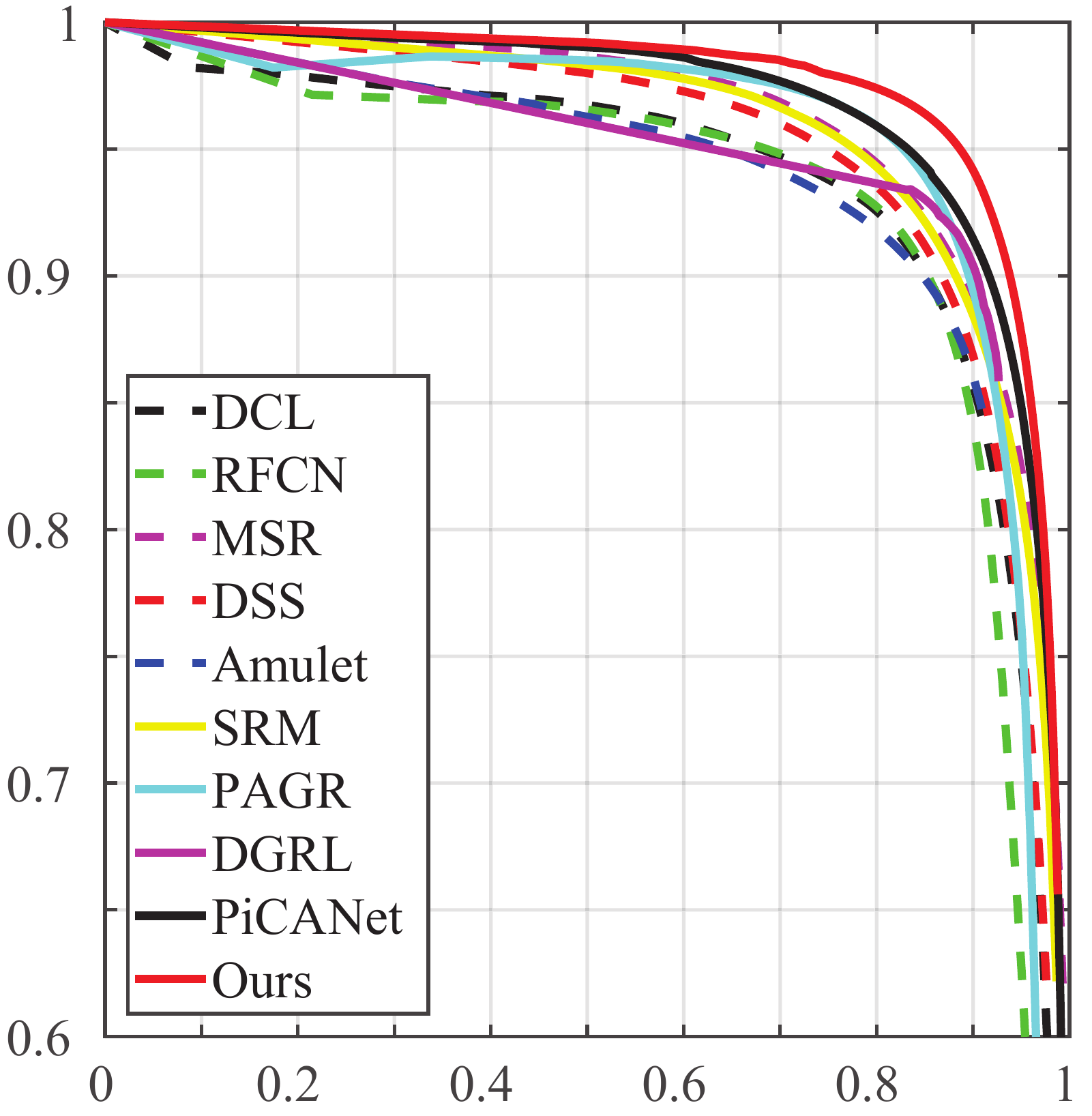} \\
		~~~~(a) DUTS-TE~\cite{wang2017learning} & ~~~~(b) DUT-OMRON~\cite{yang2013saliency} & ~~~~(c) HKU-IS~\cite{li2015visual} \\
	\end{tabular}
	\vspace{1mm}
	\caption{Precision (vertical axis) recall (horizontal axis) curves on
		three popular salient object datasets. It can be seen that the proposed method performs favorably against state-of-the-arts.}
	\label{fig:prs}
\end{figure*}
}

\vspace{-4pt}
\vspace{-4pt}
\subsubsection{The complementary information modeling} \label{sec:comple}
\vspace{-4pt}
\quad In this subsection, we explore the role of salient edge information, which is also our 
basic idea. 
The baseline is the U-Net architecture which integrates the multi-scale features (From Conv2-2 to Conv6-3) 
in the way as PSFEM (\figref{fig:arch}).  
We remove the side path $S^{(2)}$ in the baseline 
and then  
fuse the final saliency features $\hat{F}^{(3)}$ (side path from Conv3-3) and the local Conv2-2 features to 
obtain the salient edge features. Finally, we integrate salient edge features and the  salient object features 
$\hat{F}^{(3)}$ to 
get the prediction mask. 
We denote this strategy of using edges as edge\_PROG. 
The result is shown in the second row of \tabref{tab:module}. It proves that the salient edge information is 
very useful for the salient object detection task. 

\vspace{-4pt}
\subsubsection{Top-down location propagation}
\vspace{-4pt}
\quad In this subsection, we explore the role of top-down location propagation. 
Compared with edge\_PROG mentioned in the previous subsection \secref{sec:comple}, we leverage the top-down location propagation to extract more 
accurate location information from top-level instead of side path $S^{(3)}$. 
We call this strategy of using edges as edge\_TDLP.
By comparing the second and third rows of \tabref{tab:module}, the effect of top-down location propagation could be proved. 
Besides, comparing the first row and the third row of \tabref{tab:module}, we 
can find that through our explicit modeling of these two kinds of complementary information within the network, the 
performance is greatly improved on the datasets (3.1\%, 2.4\% under F-measure) without additional time and space consumption.

\renewcommand{\addFig}[1]{\includegraphics[width=0.19\linewidth]{Imgs/#1}}
\renewcommand{\addFigsd}[1]{\addFig{#1.jpg} &\addFig{#1_base.png} & \addFig{#1_nldf.png}& \addFig{#1_edge.png}& \addFig{#1.png}}
\CheckRmv{
\begin{figure}
	\centering
	\footnotesize
	\setlength\tabcolsep{0.2mm}
	\renewcommand\arraystretch{0.8}
	\begin{tabular}{ccccc}
		\addFigsd{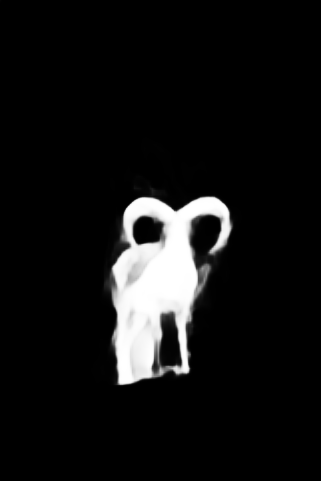}\\
		\addFigsd{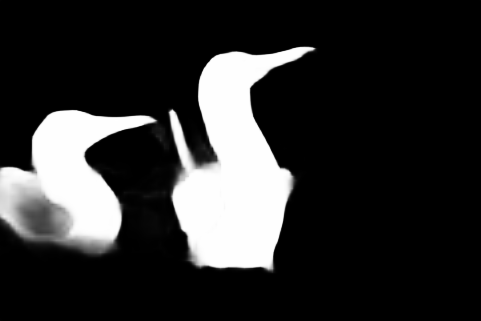}\\
		Source&B & B+edge\_NLDF &B+edge\_TDLP & GT
		\\
	\end{tabular}
	\vspace{1mm}
	\caption{Visual examples before and after adding edge cues. 
		B denotes the baseline model. edge\_NLDF and edge\_TDLP represent the edges penalty used in NLDF \cite{luo2017non} and 
		the edge modeling method proposed in this paper. The details are introduced in the \secref{sec:ablation}. 
	} 
	\label{fig:comp_nldf}
	\vspace{-5pt}
\end{figure}
}
\vspace{-4pt}

\subsubsection{Mechanism of using edge cues}
\vspace{-4pt}
\quad 
To demonstrate the advantages over NLDF \cite{luo2017non}, in which an IOU loss is added to the 
end of the network to  
punish the errors of edges.  We add the same 
IOU loss to the baseline. 
This strategy is called edge\_NLDF.
The performance is shown in the 4th row of \tabref{tab:module}. 
Compared with the baseline model, the improvement is limited. 
This also demonstrates that the proposed method of using edge information is more effective. 
The visualization results are shown in \figref{fig:comp_nldf}. Compared with 
the baseline model without edge constraint, after we add the edge penalty used in NLDF \cite{luo2017non}, edge 
information can only help refine the boundaries. 
In particular, this penalty can not help to remove the redundant parts in saliency prediction mask, nor can it 
make up for the missing parts. 
In contrast, the proposed complementary information modeling method considers the complementarity between salient edge information and salient object information, and performs better on both segmentation and localization. 

Besides, in order to further prove that salient edge detection and salient object detection are mutually helpful and complementary. 
We compare the salient edges generated by NLDF with the salient edges generated by us. 
The pre-trained model and code are both provided by the authors. 
As shown in \tabref{tab:salientedge}, it could be found that the salient edge generated by our method is much better, especially under 
the recall and F-measure metrics. 
It proves that the edges  are more accurate in our methods.

\subsubsection{The complementary features fusion}
\vspace{-4pt}
\quad After we obtain the salient edge features and multi-resolution salient object features. We aim to 
fuse these complementary features. Here we compare three fusion methods. 
The first way is the default way, which  
integrates the salient edge features ($F_E$) and the salient object features $\hat{F}^{(3)}$ which is on the top of U-Net architecture. 
The second way is to fuse the multi-resolution features $\hat{F}^{(3)}$, $\hat{F}^{(4)}$, $\hat{F}^{(5)}$, $\hat{F}^{(6)}$ 
progressively, which is called MRF\_PROG. The third way is the proposed one-to-one guidance, which is denoted MRF\_OTO. Here MRF denotes 
the multi-resolution fusion.  
The results are shown in the third, fifth, sixth rows of \tabref{tab:module}, respectively. 
It can be seen that our proposed one-to-one guidance method is most suitable for our whole architecture.

\CheckRmv{
\begin{table}[tp!] 
  \centering
  \scriptsize
  \renewcommand{\arraystretch}{1.2}
  \renewcommand{\tabcolsep}{1.5mm}
  \begin{tabular}{l| c|c|c|c|c|c  }
  	 \hline
  	 \multirow{2}{*}{Model} 
  	 & \multicolumn{3}{c|}{SOD} & \multicolumn{3}{c}{DUTS} \\
  	 \cline{2-7}
    & Recall~$\uparrow$ & Precision~$\uparrow$ & MaxF~$\uparrow$& Recall~$\uparrow$ & Precision~$\uparrow$ & MaxF~$\uparrow$ \\
\hline
NLDF& 0.513&0.541 & 0.527 &0.318 &0.659 & 0.429 \\
Ours & 0.637&0.534 & 0.581 &0.446 &0.680 & 0.539 \\
\hline
\end{tabular}
\vspace{2pt}
\caption{Comparisons on the salient edge generated by the NLDF and ours. 
} \label{tab:salientedge}
\end{table}
}

\vspace{-4pt}
\subsection{Comparison with the State-of-the-art}
\vspace{-4pt}
In this section, we compare our proposed EGNet with 15 previous state-of-the-art methods,
including DCL~\cite{li2016deep}, DSS~\cite{HouCvpr2017Dss,HouPami19Dss}, NLDF~\cite{luo2017non}, MSR~\cite{li2017instance},
ELD~\cite{lee2016deep}, DHS~\cite{liu2016dhsnet}, RFCN~\cite{wangsaliency},
  UCF~\cite{zhang2017learning},
Amulet~\cite{zhang2017amulet}, PAGR~\cite{zhang2018progressive}, PiCANet~\cite{liu2018picanet},
SRM~\cite{wang2017stagewise}, DGRL~\cite{wang2018detect}, RAS~\cite{chen2018reverse} and C2S~\cite{li2018contour}. 
Note that all the saliency maps of the above methods
are produced by running source codes or pre-computed by the authors. The 
evaluation codes are provided in \cite{HouCvpr2017Dss,HouPami19Dss,fan2017structure}.

\renewcommand{\addFig}[1]{\includegraphics[width=0.087\linewidth]{vis_new/#1}}
\newcommand{\addFigs}[1]{\addFig{#1.jpg} & \addFig{#1.png} & \addFig{#1_f.png} &
			\addFig{#1_PiCA.png} & \addFig{#1_PAGR.png} & \addFig{#1_DGRL.png} & \addFig{#1_SRM.png} 
		& \addFig{#1_UCF.png} &	\addFig{#1_AMU.png} & \addFig{#1_DSS.png} & \addFig{#1_DHS.png} }

\CheckRmv{
\begin{figure*}
	\centering
    \footnotesize
	\setlength\tabcolsep{0.2mm}
	\renewcommand\arraystretch{0.5}
	\begin{tabular}{ccccccccccc}
		\addFigs{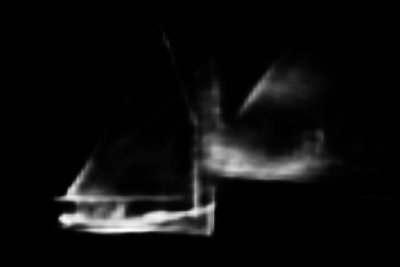}\\
		\addFigs{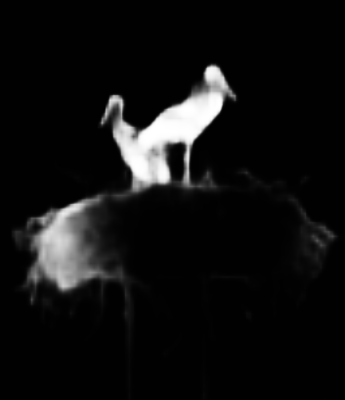}\\
		\addFigs{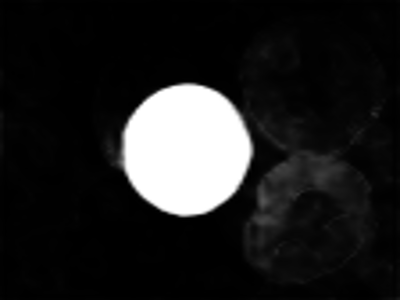}\\
		\addFigs{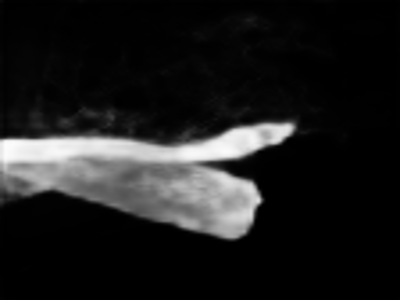}\\
		\addFigs{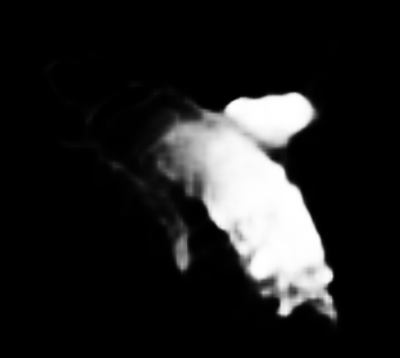}\\
		\addFigs{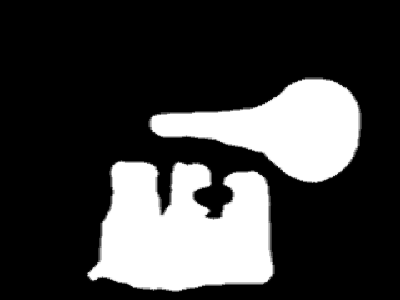}\\
		\addFigs{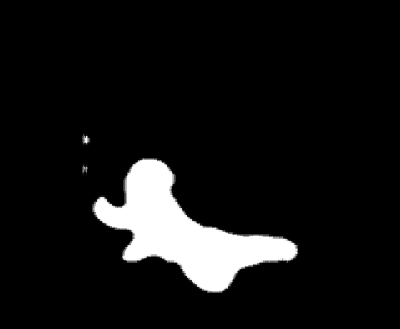}\\
	   Image & GT & Ours & PiCANet\cite{liu2018picanet} & PAGR\cite{zhang2018progressive}& DGRL\cite{wang2018detect}  &
	    SRM\cite{wang2017stagewise} & UCF\cite{zhang2017learning}& Amulet\cite{zhang2017amulet} & DSS\cite{HouCvpr2017Dss,HouPami19Dss} &
	    DHS\cite{liu2016dhsnet}\\
	\end{tabular}
	\vspace{2pt}
	\caption{Qualitative comparisons with state-of-the-arts.}
	\label{fig:vis_comps}
	\vspace{-10pt}
\end{figure*}
}

\textbf{F-measure, MAE, and S-measure.} We evaluate and compare our proposed method with other salient object detection methods in 
term of F-measure, MAE, and S-measure as shown in \tabref{tab:results}. 
We could see that different methods may use different backbone net.
Here for a fair comparison, we train our model on the VGG \cite{simonyan2014very} and ResNet \cite{he2016deep}, respectively.  
It can be seen that our model performs favorably against the state-of-the-art methods   
under all evaluation metrics  on all the compared datasets especially on the relative challenging 
dataset SOD  \cite{martin2001database, jiang2013salient} (2.9\% and 1.7\% improvements in F-measure and S-measure) and the largest dataset DUTS \cite{wang2017learning} (3.0\% and 2.5\%). 
Specifically, 
Compared with the current best approach \cite{liu2018picanet}, the average F-measure improvement  on six datasets is  
1.9\%.
Note that 
this is achieved without any pre-processing and post-processing. 

\textbf{Precision-recall curves.} Besides the numerical comparisons shown in \tabref{tab:results}, we plot the 
precision-recall curves of all compared methods over three datasets \figref{fig:prs}. 
As can be seen that the solid red line which denotes the proposed method outperforms all other methods at most thresholds. 
Due to the help of the complementary salient edge information,  the results yield sharp edge information and accurate 
localization, which results in a better PR curve. 



\textbf{Visual comparison.}
In \figref{fig:vis_comps}, we show some visualization results. 
It could be seen that our method performs better on salient object segmentation and localization. 
It is worth mentioning that thank to the salient edge features, our result could not only 
highlight the salient region but also produce coherent edges. 
For instance, for the first sample, due to the influence of the complex scene, other methods	 
are not capable of localizing and segmenting salient objects accurately. However, benefiting  from the 
complementary salient edge features, our 
method performs better. 
For the second sample, in which the salient object is relatively small, 
our result is still very close to the ground-truth.
%


\vspace{-4pt}
\section{Conclusion}
\vspace{-4pt}
In this paper, 
we aim to preserve salient object boundaries well. 
Different from other methods which integrate the multi-scale features or leverage the post-processing, 
we focus on the complementarity between salient edge information and salient object information.
Based on this idea, we propose the EGNet to model these complementary features within the network. 
First, we extract the multi-resolution salient object features based on U-Net. 
Then, we propose a non-local salient edge features extraction module which integrates the local edge information and 
global location information to get the salient edge features.
Finally, we adopt a one-to-one guidance module to fuse these complementary features.
The salient object boundaries and localization are improved under the help of salient edge features. 
 Our model 
 performs favorably against the state-of-the-art methods on six widely used datasets  without any 
 pre-processing or post-processing. 
We also provide analyses of the effectiveness of the EGNet.

\vspace{-4pt}
\paragraph{Acknowledgments.} This research was supported by NSFC (61572264), 
the national youth talent support program, 
and Tianjin Natural Science Foundation (17JCJQJC43700, 18ZXZNGX00110).

{\small
\bibliographystyle{ieee_fullname}
\bibliography{edgeS}

\begin{thebibliography}{10}\itemsep=-1pt

\bibitem{BorjiCVM2019}
Ali Borji, Ming-Ming Cheng, Qibin Hou, Huaizu Jiang, and Jia Li.
\newblock Salient object detection: A survey.
\newblock {\em CVM}, 5(2):117--150, 2019.

\bibitem{borji2015salient}
Ali Borji, Ming-Ming Cheng, Huaizu Jiang, and Jia Li.
\newblock Salient object detection: A benchmark.
\newblock {\em {IEEE TIP}}, 24(12):5706--5722, 2015.

\bibitem{chen2018reverse}
Shuhan Chen, Xiuli Tan, Ben Wang, and Xuelong Hu.
\newblock Reverse attention for salient object detection.
\newblock In {\em {ECCV}}, pages 234--250, 2018.

\bibitem{tog09Sketch2Photo}
Tao Chen, Ming-Ming Cheng, Ping Tan, Ariel Shamir, and Shi-Min Hu.
\newblock Sketch2photo: Internet image montage.
\newblock {\em {ACM TOG}}, 28(5):124:1--10, 2009.

\bibitem{cheng2015global}
Ming Cheng, Niloy~J Mitra, Xumin Huang, Philip~HS Torr, and Song Hu.
\newblock Global contrast based salient region detection.
\newblock {\em {IEEE TPAMI}}, 37(3):569--582, 2015.

\bibitem{cheng2010repfinder}
Ming-Ming Cheng, Fang-Lue Zhang, Niloy~J Mitra, Xiaolei Huang, and Shi-Min Hu.
\newblock Repfinder: finding approximately repeated scene elements for image
  editing.
\newblock {\em {ACM TOG}}, 29(4):83, 2010.

\bibitem{einhauser2003does}
Wolfgang Einh{\"a}user and Peter K{\"o}nig.
\newblock Does luminance-contrast contribute to a saliency map for overt visual
  attention?
\newblock {\em European Journal of Neuroscience}, 17(5):1089--1097, 2003.

\bibitem{everingham2010pascal}
Mark Everingham, Luc Van~Gool, Christopher~KI Williams, John Winn, and Andrew
  Zisserman.
\newblock The pascal visual object classes (voc) challenge.
\newblock {\em {IJCV}}, 88(2):303--338, 2010.

\bibitem{fan2018salient}
Deng-Ping Fan, Ming-Ming Cheng, Jiang-Jiang Liu, Shang-Hua Gao, Qibin Hou, and
  Ali Borji.
\newblock Salient objects in clutter: Bringing salient object detection to the
  foreground.
\newblock In {\em {ECCV}}, pages 186--202. Springer, 2018.

\bibitem{fan2017structure}
Deng-Ping Fan, Ming-Ming Cheng, Yun Liu, Tao Li, and Ali Borji.
\newblock Structure-measure: A new way to evaluate foreground maps.
\newblock In {\em {ICCV}}, pages 4548--4557, 2017.

\bibitem{fan2019rethinking}
Deng-Ping Fan, Zheng Lin, Jia-Xing Zhao, Yun Liu, Zhao Zhang, Qibin Hou,
  Menglong Zhu, and Ming-Ming Cheng.
\newblock Rethinking rgb-d salient object detection: Models, datasets, and
  large-scale benchmarks.
\newblock {\em arXiv preprint arXiv:1907.06781}, 2019.

\bibitem{fan2019shifting}
Deng-Ping Fan, Wenguan Wang, Ming-Ming Cheng, and Jianbing Shen.
\newblock Shifting more attention to video salient object detection.
\newblock In {\em {CVPR}}, pages 8554--8564, 2019.

\bibitem{lee2016deep}
Lee Gayoung, Tai Yu-Wing, and Kim Junmo.
\newblock Deep saliency with encoded low level distance map and high level
  features.
\newblock In {\em {CVPR}}, 2016.

\bibitem{guan2018edge}
Wenlong Guan, Tiantian Wang, Jinqing Qi, Lihe Zhang, and Huchuan Lu.
\newblock Edge-aware convolution neural network based salient object detection.
\newblock {\em IEEE SPL}, 26(1):114--118, 2018.

\bibitem{he2012mobile}
Junfeng He, Jinyuan Feng, Xianglong Liu, Tao Cheng, Tai-Hsu Lin, Hyunjin Chung,
  and Shih-Fu Chang.
\newblock Mobile product search with bag of hash bits and boundary reranking.
\newblock In {\em {CVPR}}, pages 3005--3012, 2012.

\bibitem{he2016deep}
Kaiming He, Xiangyu Zhang, Shaoqing Ren, and Jian Sun.
\newblock Deep residual learning for image recognition.
\newblock In {\em {ICCV}}, pages 770--778, 2016.

\bibitem{HouCvpr2017Dss}
Qibin Hou, Ming-Ming Cheng, Xiaowei Hu, Ali Borji, Zhuowen Tu, and Philip Torr.
\newblock Deeply supervised salient object detection with short connections.
\newblock In {\em {CVPR}}, pages 3203--3212, 2017.

\bibitem{HouPami19Dss}
Qibin Hou, Ming-Ming Cheng, Xiaowei Hu, Ali Borji, Zhuowen Tu, and Philip Torr.
\newblock Deeply supervised salient object detection with short connections.
\newblock {\em IEEE TPAMI}, 41(4):815--828, 2019.

\bibitem{hou2018selferasing}
Qibin Hou, Peng-Tao Jiang, Yunchao Wei, and Ming-Ming Cheng.
\newblock Self-erasing network for integral object attention.
\newblock In {\em {NIPS}}, 2018.

\bibitem{hu2017deep}
Ping Hu, Bing Shuai, Jun Liu, and Gang Wang.
\newblock Deep level sets for salient object detection.
\newblock In {\em {CVPR}}, pages 2300--2309, 2017.

\bibitem{itti2001computational}
Laurent Itti and Christof Koch.
\newblock Computational modeling of visual attention.
\newblock {\em Nature reviews neuroscience}, 2(3):194--203, 2001.

\bibitem{itti1998model}
Laurent Itti, Christof Koch, and Ernst Niebur.
\newblock A model of saliency-based visual attention for rapid scene analysis.
\newblock {\em {IEEE TPAMI}}, 20(11):1254--1259, 1998.

\bibitem{jia2019richer}
Sen Jia and Neil~DB Bruce.
\newblock Richer and deeper supervision network for salient object detection.
\newblock {\em arXiv preprint arXiv:1901.02425}, 2019.

\bibitem{klein2011center}
Dominik~A Klein and Simone Frintrop.
\newblock Center-surround divergence of feature statistics for salient object
  detection.
\newblock In {\em {ICCV}}, pages 2214--2219. IEEE, 2011.

\bibitem{lecun1998gradient}
Yann LeCun, L{\'e}on Bottou, Yoshua Bengio, and Patrick Haffner.
\newblock Gradient-based learning applied to document recognition.
\newblock {\em Proceedings of the IEEE}, 86(11):2278--2324, 1998.

\bibitem{li2017instance}
Guanbin Li, Yuan Xie, Liang Lin, and Yizhou Yu.
\newblock Instance-level salient object segmentation.
\newblock In {\em {CVPR}}, 2017.

\bibitem{li2015visual}
Guanbin Li and Yizhou Yu.
\newblock Visual saliency based on multiscale deep features.
\newblock In {\em {CVPR}}, pages 5455--5463, 2015.

\bibitem{li2016deep}
Guanbin Li and Yizhou Yu.
\newblock Deep contrast learning for salient object detection.
\newblock In {\em {CVPR}}, 2016.

\bibitem{li2018contour}
Xin Li, Fan Yang, Hong Cheng, Wei Liu, and Dinggang Shen.
\newblock Contour knowledge transfer for salient object detection.
\newblock In {\em {ECCV}}, pages 355--370, 2018.

\bibitem{li2014secrets}
Yin Li, Xiaodi Hou, Christof Koch, James~M Rehg, and Alan~L Yuille.
\newblock The secrets of salient object segmentation.
\newblock In {\em {CVPR}}, pages 280--287, 2014.

\bibitem{li2019deep}
Zun Li, Congyan Lang, Yunpeng Chen, Junhao Liew, and Jiashi Feng.
\newblock Deep reasoning with multi-scale context for salient object detection.
\newblock {\em arXiv preprint arXiv:1901.08362}, 2019.

\bibitem{liu2016dhsnet}
Nian Liu and Junwei Han.
\newblock Dhsnet: Deep hierarchical saliency network for salient object
  detection.
\newblock In {\em {CVPR}}, pages 678--686, 2016.

\bibitem{liu2018picanet}
Nian Liu, Junwei Han, and Ming-Hsuan Yang.
\newblock Picanet: Learning pixel-wise contextual attention for saliency
  detection.
\newblock In {\em {CVPR}}, pages 3089--3098, 2018.

\bibitem{long2015fully}
Jonathan Long, Evan Shelhamer, and Trevor Darrell.
\newblock Fully convolutional networks for semantic segmentation.
\newblock In {\em {CVPR}}, pages 3431--3440, 2015.

\bibitem{luo2017non}
Zhiming Luo, Akshaya~Kumar Mishra, Andrew Achkar, Justin~A Eichel, Shaozi Li,
  and Pierre-Marc Jodoin.
\newblock Non-local deep features for salient object detection.
\newblock In {\em {CVPR}}, 2017.

\bibitem{martin2001database}
David Martin, Charless Fowlkes, Doron Tal, and Jitendra Malik.
\newblock A database of human segmented natural images and its application to
  evaluating segmentation algorithms and measuring ecological statistics.
\newblock In {\em {ICCV}}, volume~2, pages 416--423, 2001.

\bibitem{movahedi2010design}
Vida Movahedi and James~H Elder.
\newblock Design and perceptual validation of performance measures for salient
  object segmentation.
\newblock In {\em {IEEE CVPRW}}, pages 49--56. IEEE, 2010.

\bibitem{mumford1989optimal}
David Mumford and Jayant Shah.
\newblock Optimal approximations by piecewise smooth functions and associated
  variational problems.
\newblock {\em CPAM}, 42(5):577--685, 1989.

\bibitem{parkhurst2002modeling}
Derrick Parkhurst, Klinton Law, and Ernst Niebur.
\newblock Modeling the role of salience in the allocation of overt visual
  attention.
\newblock {\em Vision research}, 42(1):107--123, 2002.

\bibitem{ronneberger2015u}
Olaf Ronneberger, Philipp Fischer, and Thomas Brox.
\newblock U-net: Convolutional networks for biomedical image segmentation.
\newblock In {\em International Conference on Medical image computing and
  computer-assisted intervention}, pages 234--241. Springer, 2015.

\bibitem{rosin2013artistic}
Paul~L Rosin and Yu-Kun Lai.
\newblock Artistic minimal rendering with lines and blocks.
\newblock {\em Graphical Models}, 75(4):208--229, 2013.

\bibitem{rutishauser2004bottom}
Ueli Rutishauser, Dirk Walther, Christof Koch, and Pietro Perona.
\newblock Is bottom-up attention useful for object recognition?
\newblock In {\em {CVPR}}, 2004.

\bibitem{simonyan2014very}
Karen Simonyan and Andrew Zisserman.
\newblock Very deep convolutional networks for large-scale image recognition.
\newblock In {\em ICLR}, 2015.

\bibitem{jiang2013salient}
Jingdong Wang, Huaizu Jiang, Zejian Yuan, Ming-Ming Cheng, Xiaowei Hu, and
  Nanning Zheng.
\newblock Salient object detection: A discriminative regional feature
  integration approach.
\newblock {\em {IJCV}}, 123(2):251--268, 2017.

\bibitem{wang2015deep}
Lijun Wang, Huchuan Lu, Xiang Ruan, and Ming-Hsuan Yang.
\newblock Deep networks for saliency detection via local estimation and global
  search.
\newblock In {\em {ICCV}}, pages 3183--3192, 2015.

\bibitem{wang2017learning}
Lijun Wang, Huchuan Lu, Yifan Wang, Mengyang Feng, Dong Wang, Baocai Yin, and
  Xiang Ruan.
\newblock Learning to detect salient objects with image-level supervision.
\newblock In {\em {CVPR}}, pages 136--145, 2017.

\bibitem{wang2016saliency}
Linzhao Wang, Lijun Wang, Huchuan Lu, Pingping Zhang, and Xiang Ruan.
\newblock Saliency detection with recurrent fully convolutional networks.
\newblock In {\em {ECCV}}, pages 825--841. Springer, 2016.

\bibitem{wangsaliency}
Linzhao Wang, Lijun Wang, Huchuan Lu, Pingping Zhang, and Xiang Ruan.
\newblock Saliency detection with recurrent fully convolutional networks.
\newblock In {\em {ECCV}}, 2016.

\bibitem{wang2017stagewise}
Tiantian Wang, Ali Borji, Lihe Zhang, Pingping Zhang, and Huchuan Lu.
\newblock A stagewise refinement model for detecting salient objects in images.
\newblock In {\em {ICCV}}, pages 4019--4028, 2017.

\bibitem{wangiccv2019lfsd}
Tiantian Wang, Yongri Piao, Li Xiao, Lihe Zhang, and Huchuan Lu.
\newblock Deep learning for light field saliency detection.
\newblock In {\em {ICCV}}, 2019.

\bibitem{WangECCV2016ksr}
Tiantian Wang, Lihe Zhang, Huchuan Lu, Chong Sun, and Jinqing Qi.
\newblock Kernelized subspace ranking for saliency detection.
\newblock In {\em {ECCV}}, pages 450--466, 2016.

\bibitem{wang2018detect}
Tiantian Wang, Lihe Zhang, Shuo Wang, Huchuan Lu, Gang Yang, Xiang Ruan, and
  Ali Borji.
\newblock Detect globally, refine locally: A novel approach to saliency
  detection.
\newblock In {\em {CVPR}}, pages 3127--3135, 2018.

\bibitem{wang2018salient}
Wenguan Wang, Jianbing Shen, Xingping Dong, and Ali Borji.
\newblock Salient object detection driven by fixation prediction.
\newblock In {\em {ICCV}}, pages 1711--1720, 2018.

\bibitem{RANet2019}
Ziqin Wang, Jun Xu, Li Liu, Fan Zhu, and Ling Shao.
\newblock Ranet: Ranking attention network for fast video object segmentation.
\newblock In {\em {ICCV}}, Oct 2019.

\bibitem{xie2015holistically}
Saining Xie and Zhuowen Tu.
\newblock Holistically-nested edge detection.
\newblock In {\em {ICCV}}, pages 1395--1403, 2015.

\bibitem{yan2013hierarchical}
Qiong Yan, Li Xu, Jianping Shi, and Jiaya Jia.
\newblock Hierarchical saliency detection.
\newblock In {\em {CVPR}}, pages 1155--1162, 2013.

\bibitem{yang2013saliency}
Chuan Yang, Lihe Zhang, Huchuan Lu, Xiang Ruan, and Ming-Hsuan Yang.
\newblock Saliency detection via graph-based manifold ranking.
\newblock In {\em {CVPR}}, pages 3166--3173, 2013.

\bibitem{zhang2017deep}
Jing Zhang, Yuchao Dai, Fatih Porikli, and Mingyi He.
\newblock Deep edge-aware saliency detection.
\newblock {\em arXiv preprint arXiv:1708.04366}, 2017.

\bibitem{zhang2018bi}
Lu Zhang, Ju Dai, Huchuan Lu, You He, and Gang Wang.
\newblock A bi-directional message passing model for salient object detection.
\newblock In {\em {ICCV}}, pages 1741--1750, 2018.

\bibitem{zhang2019salient}
Pingping Zhang, Wei Liu, Huchuan Lu, and Chunhua Shen.
\newblock Salient object detection with lossless feature reflection and
  weighted structural loss.
\newblock {\em {IEEE TIP}}, 2019.

\bibitem{zhang2017amulet}
Pingping Zhang, Dong Wang, Huchuan Lu, Hongyu Wang, and Xiang Ruan.
\newblock Amulet: Aggregating multi-level convolutional features for salient
  object detection.
\newblock In {\em {ICCV}}, pages 202--211, 2017.

\bibitem{zhang2017learning}
Pingping Zhang, Dong Wang, Huchuan Lu, Hongyu Wang, and Baocai Yin.
\newblock Learning uncertain convolutional features for accurate saliency
  detection.
\newblock In {\em {ICCV}}, pages 212--221. IEEE, 2017.

\bibitem{zhang2018progressive}
Xiaoning Zhang, Tiantian Wang, Jinqing Qi, Huchuan Lu, and Gang Wang.
\newblock Progressive attention guided recurrent network for salient object
  detection.
\newblock In {\em {CVPR}}, pages 714--722, 2018.

\bibitem{ZhaoCvm2018flic}
Jiaxing Zhao, Ren Bo, Qibin Hou, Ming-Ming Cheng, and Paul Rosin.
\newblock Flic: Fast linear iterative clustering with active search.
\newblock {\em CVM}, 4(4):333--348, Dec 2018.

\bibitem{zhao2018flic}
Jiaxing Zhao, Bo Ren, Qibin Hou, and Ming-Ming Cheng.
\newblock Flic: Fast linear iterative clustering with active search.
\newblock In {\em AAAI}, 2018.

\bibitem{zhao2019contrast}
Jia-Xing Zhao, Yang Cao, Deng-Ping Fan, Ming-Ming Cheng, Xuan-Yi Li, and Le
  Zhang.
\newblock Contrast prior and fluid pyramid integration for rgbd salient object
  detection.
\newblock In {\em {CVPR}}, 2019.

\bibitem{zhao2019optimizing}
Kai Zhao, Shanghua Gao, Wenguan Wang, and Ming-Ming Cheng.
\newblock Optimizing the f-measure for threshold-free salient object detection.
\newblock In {\em {ICCV}}, Oct 2019.

\bibitem{zhu2014saliency}
Wangjiang Zhu, Shuang Liang, Yichen Wei, and Jian Sun.
\newblock Saliency optimization from robust background detection.
\newblock In {\em {CVPR}}, pages 2814--2821, 2014.

\bibitem{zhuge2018boundary}
Yunzhi Zhuge, Gang Yang, Pingping Zhang, and Huchuan Lu.
\newblock Boundary-guided feature aggregation network for salient object
  detection.
\newblock {\em IEEE SPL}, 25(12):1800--1804, 2018.

\end{thebibliography}
}

\end{document}